\documentclass[sigconf]{acmart}

\AtBeginDocument{%
  \providecommand\BibTeX{{%
    \normalfont B\kern-0.5em{\scshape i\kern-0.25em b}\kern-0.8em\TeX}}}

\copyrightyear{2024}
\acmYear{2024}
\setcopyright{acmlicensed}
\acmConference[SIGIR '24] {Proceedings of the 47th International ACM SIGIR Conference on Research and Development in Information Retrieval}{July 14--18, 2024}{Washington, DC, USA.}
\acmBooktitle{Proceedings of the 47th International ACM SIGIR Conference on Research and Development in Information Retrieval (SIGIR '24), July 14--18, 2024, Washington, DC, USA}
\acmISBN{979-8-4007-0431-4/24/07}
\acmDOI{10.1145/3626772.3657785}

\usepackage{multirow}
\usepackage{enumitem}

\usepackage[normalem]{ulem}
\useunder{\uline}{\ul}{}

\newcommand{\eg}{\emph{e.g., }}

\newcommand{\wjc}[1]{{\color{black}{#1}}}
\newcommand{\zzh}[1]{{\color{black}{#1}}}



\author{Zihao Zhao}
\orcid{0009-0005-8074-6500}
\affiliation{%
    \institution{University of Science and Technology of China}
    \streetaddress{No.100 Fuxing Road}
    \city{Hefei}
    \state{Anhui}
    \postcode{230093}
    \country{China}
}
\email{zzh1998@mail.ustc.edu.cn}

\author{Yi Jing}
\orcid{0009-0003-5299-7471}
\affiliation{%
    \institution{University of Science and Technology of China}
    \streetaddress{96, Jinzhai Road}
    \city{Hefei}
    \state{Anhui}
    \postcode{230026}
    \country{China}
}
\email{jingi@mail.ustc.edu.cn}

\author{Fuli Feng}
\orcid{0000-0002-5828-9842}
\authornote{Corresponding authors.}
\affiliation{%
    \institution{University of Science and Technology of China}
    \streetaddress{No.100 Fuxing Road}
    \city{Hefei}
    \state{Anhui}
    \postcode{230093}
    \country{China}
}
\email{fulifeng93@gmail.com}

\author{Jiancan Wu}
\orcid{0000-0002-6941-5218}
\affiliation{%
    \institution{University of Science and Technology of China}
    \streetaddress{No.100 Fuxing Road}
    \city{Hefei}
    \state{Anhui}
    \postcode{230093}
    \country{China}
}
\email{wujcan@gmail.com}

\author{Chongming Gao}
\orcid{0000-0002-5187-9196}
\affiliation{%
    \institution{University of Science and Technology of China}
    \streetaddress{No.100 Fuxing Road}
    \city{Hefei}
    \state{Anhui}
    \postcode{230093}
    \country{China}
}
\email{chongming.gao@gmail.com}

\author{Xiangnan He}
\authornotemark[1]
\orcid{0000-0001-8472-7992}
\affiliation{%
    \institution{University of Science and Technology of China}
    \streetaddress{No.100 Fuxing Road}
    \city{Hefei}
    \state{Anhui}
    \postcode{230093}
    \country{China}
}
\email{xiangnanhe@gmail.com}

\renewcommand{\shortauthors}{Zihao Zhao, et al.}

\begin{document}

\title{Leave No Patient Behind: Enhancing Medication Recommendation for Rare Disease Patients}

\begin{abstract}
Medication recommendation systems have gained significant attention in healthcare as a means of providing tailored and effective drug combinations based on patients' clinical information. However, existing approaches often suffer from fairness issues, as recommendations tend to be more accurate for patients with common diseases compared to those with rare conditions. In this paper, we propose a novel model called Robust and Accurate REcommendations for Medication (RAREMed), which leverages the pretrain-finetune learning paradigm to enhance accuracy for rare diseases. RAREMed employs a transformer encoder with a unified input sequence approach to capture complex relationships among disease and procedure codes. Additionally, it introduces two self-supervised pre-training tasks, namely Sequence Matching Prediction (SMP) and Self Reconstruction (SR), to learn specialized medication needs and interrelations among clinical codes. Experimental results on two real-world datasets demonstrate that RAREMed provides accurate drug sets for both rare and common disease patients, thereby mitigating unfairness in medication recommendation systems. 
The implementation is available via \url{https://github.com/zzhUSTC2016/RAREMed}

\end{abstract}
\maketitle
\section{Introduction}

\begin{figure}[t!]
    \centering
    \includegraphics[width=0.5\textwidth]{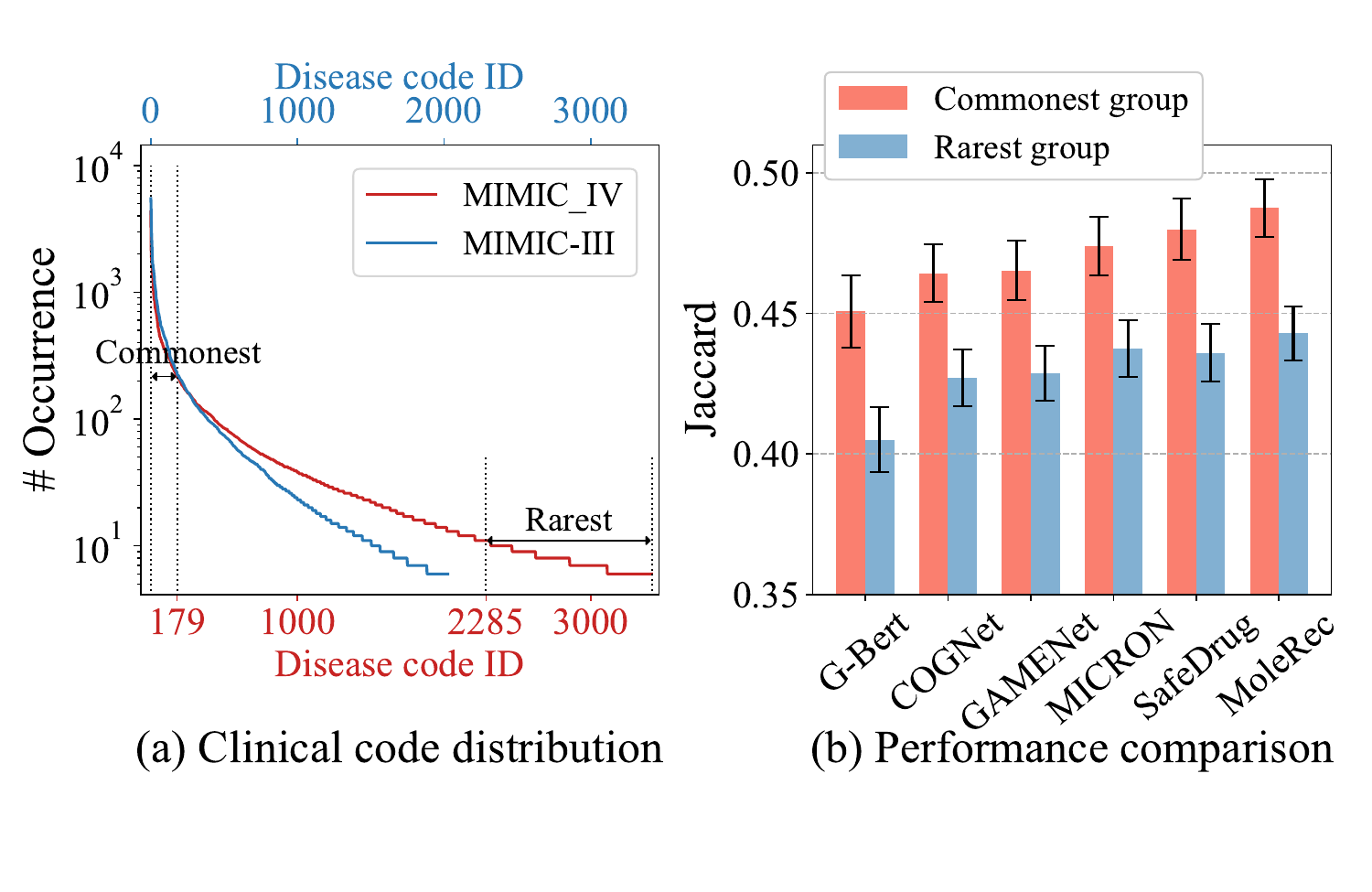}
    \vspace{-1.27cm}
    \caption{
    (a) The long-tail distribution of disease  codes in the MIMIC-III~\cite{mimic-iii} and MIMIC-IV~\cite{mimic-iv} datasets. (b) Patients in MIMIC-IV are divided into five equal-sized groups based on the ranking of their least common disease codes in terms of popularity. The subgraph illustrates the recommendation accuracy, as measured by the Jaccard index, of SOTA methods for both the commonest and the rarest code groups.}
    \label{fig:intro}
    \vspace{-0.45cm}
\end{figure}

Medication recommendation has garnered growing interest as a means of offering secure and \wjc{effective} drug combinations \wjc{tailored to} patients' clinical information~\cite{medrec-survey2023}. Through the utilization of \wjc{medical} knowledge and extensive patient data, a medication recommendation system can facilitate healthcare professionals in \wjc{prescribing} more precise and effective drug combinations while \wjc{reducing} the occurrences of medication errors and \wjc{adverse} \zzh{drug-drug interactions (DDI) among selected medicines~\cite{gamenet, safedrug}}. The substantial harm caused by inaccurate clinical decisions has led to  a profound impact on numerous patients~\cite{leap}.
\wjc{Notably, it is estimated that at least 1.5 million individuals~\cite{institute2007preventing} are harmed by preventable medication errors each year, incurring a substantial raises of healthcare expenses~\cite{cost1998}.}
Consequently, it is imperative to explore improved methodologies for medication recommendations.

\wjc{Current approaches to medication recommendation primarily focus on maximizing overall accuracy by incorporating patients' current medical conditions~\cite{leap, compnet} or clinical information from previous visits~\cite{g-bert, gamenet, safedrug, cognet, micron, drugrec}.
While significant progress has been made in comprehensive utilization these data, existing approaches encounter fairness issues in terms of recommendations across patient groups.
This stems from the highly skewed distribution of clinical codes in Electronic Health Records (EHR)~\cite{ehr} --- a digital compilation of patients' medical conditions, including diseases, procedures, medications and other clinical data during patient visits.
As illustrated in Figure~\ref{fig:intro}(a), a small subset of codes exhibit high prevalence, while the long tail comprises low-occurrence diseases in two real-world EHR datasets.
Consequently, patients resided in \zzh{the most} common group receive significantly more accurate recommendations compared to those in \zzh{the rarest} group, as depicted in Figure~\ref{fig:intro}(b).
This disparity in accuracy across patient groups \zzh{raises concerns regarding equity and fairness, ultimately undermining} the reliability and utility of medication recommendation systems.

}

\zzh{The key to promoting equity in} medication recommendation is \zzh{to improve} predictive accuracy for underserved patient populations, particularly those with rare conditions.
Nevertheless, accurately modeling and recommending medications for rare diseases remains as a challenging task. We summarize three central difficulties:
\begin{itemize}[leftmargin=*]
    \item \textbf{Scarce quality data}: 
    \zzh{Due to the inherently low prevalence of rare diseases, there is a paucity of clinical data related to these conditions in EHR~\cite{mitani2020small}. This scarcity of high-quality data severely impedes the accurate discernment of medication recommendation patterns for rare diseases.}

    \item \textbf{Intricate disease/procedure associations}\footnotemark[1]: Distinct from common diseases, rare diseases often \zzh{present} complex combinations of symptoms and clinical events~\cite{mueller2016characteristics, field2010profile}, which confounds efforts to \zzh{elucidate} reliable disease associations. \zzh{Moreover}, the complex relationships between diseases and procedures \zzh{further obfuscate appropriate medication selection}~\cite{medical_example1, medical_example2} --- certain treatments may be commonly applied across multiple conditions while the \zzh{suitability} varies depending on disease severity.

    \item \textbf{Tailored medication needs}\footnote[1]{We will further discuss the profile of rare disease patients in Section~\ref{apd:difficulty}.}: 
    \zzh{The atypical pathophysiology of rare diseases often necessitates more precisely targeted pharmacological interventions. On the other hand, patients with rare diseases tend to have more complicated therapeutic regimens, warrant tailored medication recommendations.}

\end{itemize}

In this work, we aim to foster fair medication recommendations by uplifting their accuracy on rare diseases.
We resort to the pretrain-finetune learning paradigm for addressing the aforementioned challenges, drawing inspiration from its success in natural language processing~\cite{bert, gpt} and computer vision~\cite{resnet, chen2020simple}, which yet remains largely unexplored in medication recommendation.
Compared to traditional learning paradigms, \eg supervised learning, pretrain-finetune enjoys the merit of exploiting the unlabeled data space through self-supervised pretraining, showing great potential in modeling intricate relations and addressing data scarcity.
However, implementing pretrain-finetune in fair medication recommendations presents challenges, which we summarize in two key questions:
(1) \textbf{How to build an expressive encoder to fully capture patients' clinical information?}
This encoder should be able to handle the complexity of rare diseases, ensuring that the medication recommendations are based on a thorough understanding of the patient's condition.
(2) \textbf{How to design optimal pretraining objectives to learn rare diseases' specialized medication needs and interrelations among diseases, procedures, and medications?}

These two questions motivate us to propose a novel model called Robust and Accurate REcommendations for Medication (RAREMed), which brings the superiority of pretrain-finetune to representation learning for fair medication recommendations.
RAREMed utilizes a transformer encoder architecture, which is effective at capturing complex relationships.
To comprehensively represent patient information, unlike prior studies that overlook the intricate connections between diseases and procedures~\cite{gamenet, safedrug, drugrec, cognet, molerec}, we introduce a unified sequence approach to jointly encode both components. This allows modeling potential associations, \eg certain procedures \zzh{are} applied differently depending on disease type and severity\footnote[2]{
For example, the typical choice of non-invasive shock wave lithotripsy for kidney stones suggests a better overall health condition for the patient, whereas endoscopic surgery is considered a more severe option.
~\cite{medical_example2}.
In contrast, prostate enlargement is commonly treated with endoscopic surgery for mild cases and shock wave lithotripsy for severe cases involving prostate stones~\cite{medical_example1}.}~\cite{medical_example1,medical_example2}, better capturing rare disease complexity for accurate recommendations.
Additionally, we enrich the standard token embedding layer with \textit{segment embeddings} and \textit{relevance embeddings}, where segment embeddings help differentiate between disease and procedure codes, while relevance embeddings capture their varying importance. 
To learn rare diseases' tailored needs and maximize the utilization of available data, we design two self-supervised pretraining tasks: Sequence Matching Prediction (SMP) and Self Reconstruction (SR).
SMP predicts whether a disease-procedure sequence pair belongs to the same patient, enabling the learning of contextual dependencies and underlying connections \zzh{among clinical codes.}
While SR reconstructs the input sequence from the patient representation, promoting a comprehensive comprehension of all codes, especially the rare ones.
By pretraining on these objectives, RAREMed acquires robust encoding abilities, even for limited clinical codes.
Through experiments on two public real-world datasets, we demonstrate that RAREMed provides accurate drug sets for both rare disease and common disease patients, mitigating the unfairness in the medication recommendation system.

In summary, our work makes the following main contributions:
\begin{itemize}[leftmargin=*]
\item To our knowledge, this is the first work to address the unfairness issue in medication recommendation, whereby patients with rare diseases cannot obtain accurate recommendations.

\item We propose a novel medication recommendation model, RAREMed, which combines pre-training techniques to learn robust representations for rare diseases, thereby improving the accuracy and mitigating unfairness in medication recommendation.

\item Extensive experiments on two benchmark datasets demonstrate RAREMed's superiority over a range of state-of-the-art methods. 
We will release the source code to facilitate future research.

\end{itemize}

\section{Related Works}

\noindent\textbf{Medication Recommendation.}
Methods for medication recommendation can be broadly categorized into two groups: longitudinal and instance-based approaches.  Longitudinal methods make use of patients' longitudinal medical history, such as the approach taken by Choi et al.~\cite{retain}, who employ a two-level temporal attention mechanism. Similarly, Shang et al.~\cite{g-bert} pre-train their model on single-visit data\footnote[3]{Single-visit patients have no historical visits, i.e., have only one visit recorded in EHR.} and fine-tune it using multi-visit data\footnote[4]{Multi-visit patients have at least two visits in EHR.}. Additionally, Shang et al.~\cite{gamenet} incorporate a graph-augmented memory module and DDI graph to reduce adverse drug-drug interactions, while Yang et al.~\cite{safedrug} consider drug molecule structures for medication security. \citet{cognet} introduce a copy-or-predict mechanism, and Yang et al.~\cite{micron} propose a residual-based recurrent network. Sun et al.~\cite{drugrec} present a causal model to address recommendation bias, and Yang et al.~\cite{molerec} leverage a molecular substructure-aware attentive method. More recently, Bhoi et al.~\cite{refine} propose a fine-grained medication recommendation approach. However, many of these methods heavily rely on historical records, making them less suitable for single-visit patients~\cite{retain, g-bert, drugrec, refine}, or exhibit reduced accuracy in such cases~\cite{gamenet, safedrug, cognet, micron, molerec}.
Instance-based methods focus on the current medical condition of patients. For instance, Zhang et al.~\cite{leap} 
utilize an RNN model based on disease codes. Wang et al.~\cite{compnet} 
employ Deep Q Learning to capture correlations and adverse interactions between medicines. However, these models may not adequately capture patient-specific information, potentially leading to reduced accuracy.
Despite the initial success of these methods, a common issue among them is fairness~\cite{bias-survey, fairness-survey}. Patients with common diseases tend to receive more accurate medication predictions compared to those with rare diseases. This discrepancy undermines the overall performance and impairs the reliability and usefulness of medication recommendation systems.

\noindent\textbf{Fairness in Recommendation.}
Fairness concerns in recommender systems have garnered significant attention due to observed systematic and unfair discrimination against certain individuals or groups~\cite{fairness-survey, bias-survey}. Approaches to achieve fairness in machine learning can be categorized into three groups: Pre-processing, In-processing, and Post-processing~\cite{fairness-survey2}.
Pre-processing methods modify the training data before the learning process to eliminate biases. For example, Ekstrand et al.~\cite{ekstrand2018all} address gender bias in movie/music recommendation by creating gender-balanced data through random resampling. Rastegarpanah et al.~\cite{rastegarpanah2019fighting} propose adding "antidote" data to improve the social desirability of recommender system outputs. However, in medication recommendation, it is ethically unfeasible to modify individual patients' Electronic Health Record (EHR) data, limiting the effectiveness of these pre-processing methods.
In-processing methods focus on removing sensitive attribute information through regularization or adversarial learning. For instance, Yao et al.~\cite{yao2017beyond} propose fairness metrics as regularization in collaborative filtering. Li et al.~\cite{li2021leave} introduce a text-based reconstruction loss to ensure more balanced recommendation utility across all users. However, these methods often involve non-convex optimization and the fairness issues in this research is irrelevant to sensitive attributes.
Post-processing methods involve re-ranking the output list of base ranking models to improve fairness. For instance, Geyik et al.~\cite{geyik2019fairness} employ interval constrained ranking to achieve multiple group fairness. However, in medication recommendation, the focus is on recommending a set of drugs rather than a ranking of individual drugs, making the relevance of re-ranking less significant in this task.
In short, while methods have been proposed to tackle unfairness in recommender systems, their applicability to the specific fairness challenges in medication recommendation is limited.

\noindent\textbf{Pre-training in Recommendation.}
Pre-training techniques are widely used in feature-based recommendation systems to enhance user or item representations~\cite{pretrain_survey}.  Qiu et al.~\cite{qiu2021u} propose review encoder pre-training to complement user representations, while Wong et al.~\cite{wong2021improving} utilize pre-training on a large-scale knowledge graph for conversational recommender systems. However, these methods are not applicable to medication recommendation. Shang et al.~\cite{g-bert} introduce a pre-training method to address selection bias, but it lacks procedure coding and recommendations for patients without historical records, and fairness concerns remain unresolved.
\section{Problem Formulation}
\label{sec:formulation}
\noindent\textbf{Electronic Health Record (EHR).}
An EHR, denoted as $\mathcal{R} = {\{\mathcal{V}^{(j)}\}_{j=1}^N}$, represents a collection of medical records for N patients. In this context, $\mathcal{D} = \{d_1, d_2, \cdots, d_{|\mathcal{D}|} \}$ refers to the set of diseases, $\mathcal P = \{p_1, p_2, \cdots, p_{|\mathcal{P}|}\}$ denotes the set of procedures, and $\mathcal M = \{m_1, m_2, \cdots, m_{|\mathcal{M}|}\}$ represents the set of medications. The notation $|\cdot|$ indicates the cardinality of a set.

The record of patient $j$, denoted as $\mathcal{V}^{(j)} = \{\mathbf{d}^{(j)}, \mathbf{p}^{(j)}, \mathbf{m}^{(j)}\}$, comprises three components. Firstly, $\mathbf{d}^{(j)} = [d_1, d_2, \cdots, d_x] \in \mathcal{D}$ represents the sequence of diseases experienced by the patient, ordered by their priority or significance\footnote[5]{The sequence numbers of priority are provided in the EHR datasets~\cite{mimic-iii, mimic-iv}, which are labeled by qualified physicians.}, reflecting their relative importance in the patient's healthcare journey. Similarly, $\mathbf{p}^{(j)}  = [p_1, p_2, \cdots, p_y] \in \mathcal{P}$ captures the sequence of procedures, also ordered by priority. Here, $x$ and $y$ denote the lengths of the disease and procedure sequences, respectively. 
Lastly, $\mathbf{m}^{(j)} \in \{0, 1\}^{|\mathcal{M}|}$ is a multi-hot vector indicating the medications prescribed to the patient. Each element in $\mathbf{m}^{(j)}$ corresponds to a specific medication in $\mathcal{M}$, and a value of 1 indicates that the medication was prescribed, while 0 indicates its absence. We will simplify the notation by omitting the superscript $j$ whenever there is no ambiguity.

\noindent\textbf{DDI Graph.}
The Drug-Drug Interaction (DDI) Graph~\cite{gamenet, safedrug, ddi_review2015} is a tool used to assess and mitigate the risk of adverse drug-drug interactions within recommended medication combinations. It is represented by a symmetric binary adjacency matrix $\mathbf{A} \in \{0, 1\}^{|\mathcal M|\times|\mathcal M|}$, where each element $\mathbf{A}_{uv}$ is assigned a value of 1 only if there exists a harmful interaction between the $u$-th and $v$-th drugs.

\noindent\textbf{Medication Recommendation Problem.}
Given a patient's current disease sequence $\mathbf{d}$ and procedure sequence $\mathbf{p}$, along with the DDI graph A, the objective is to generate an appropriate combination of medications $\hat{\mathcal{Y}}$ for the patient that maximizes prediction accuracy while minimizing the risk of adverse drug interactions.

\noindent\textbf{Fair Medication Recommendation Problem.}
In addition to the conventional focus on overall accuracy, our research emphasizes the fairness of medication recommendation. Fairness in this context refers to providing accurate recommendations for both patients with common diseases and those with rare diseases. 
\section{Profile of Rare Disease Patients}
\label{apd:difficulty}

\begin{figure}[t]
\centering
\includegraphics[width=0.47\textwidth]{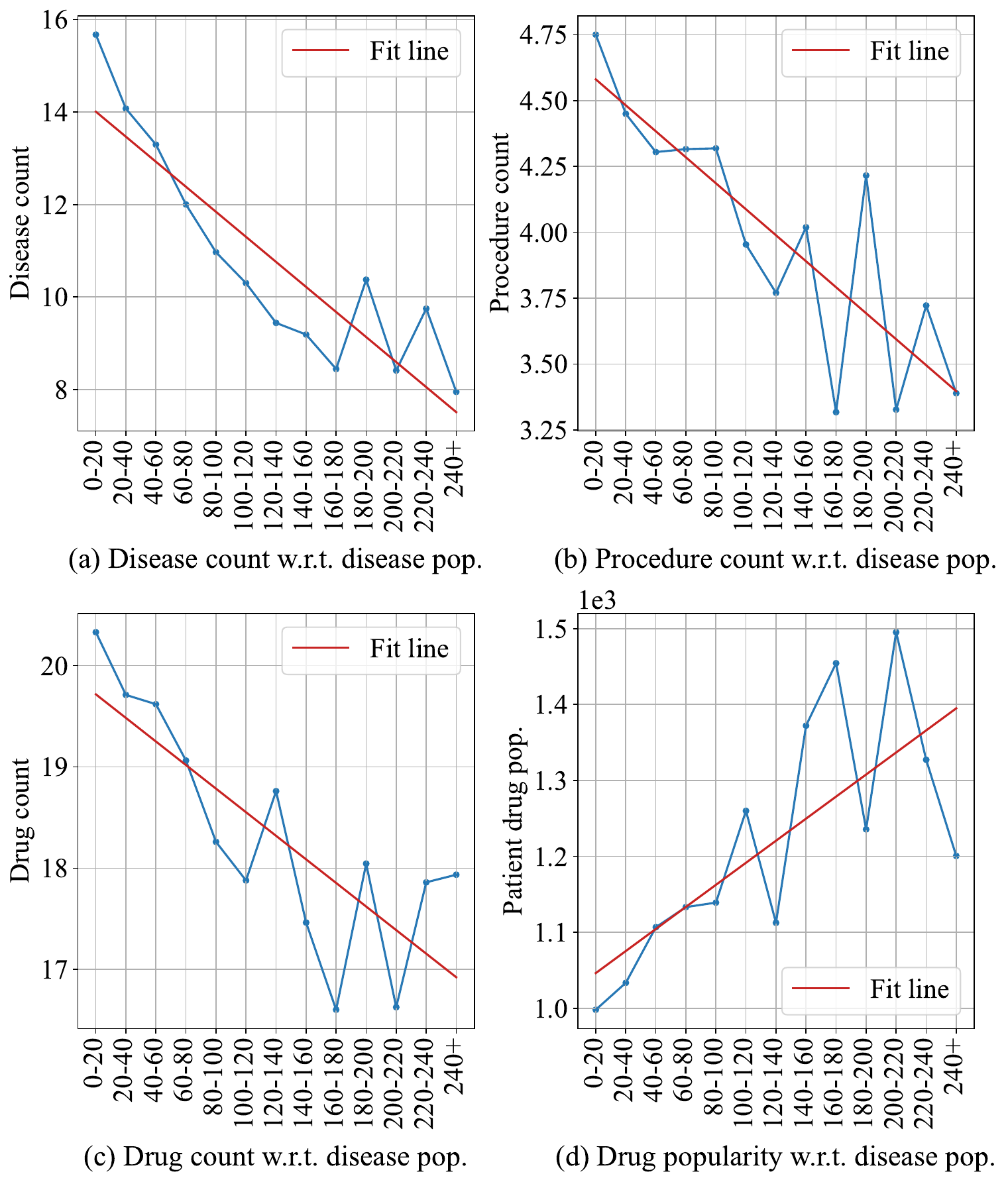}
\vspace{-0.2cm}
\caption{
We categorize patients into 13 distinct groups based on the frequency of their least common disease codes. Subplots (a), (b), (c), and (d) depict the relationship between the average disease count, procedure count, drug count, and drug popularity, respectively, \zzh{w.r.t.} the rarest disease popularity.
}
\vspace{-0.4cm}

\label{fig:difficulty}
\end{figure}

In this section, we conduct empirical analysis on real-world EHR dataset MIMIC-IV~\cite{mimic-iv} to explore the profile of patients with rare diseases. 
Specifically, we categorize patients into 13 groups based on the popularity of their rarest disease codes, using uniform segmentation across popularity intervals. We then calculate various metrics
for each group. 
From the results presented in Figure~\ref{fig:difficulty}, we make two important observations:

\textbf{Observation 1}: The data presented in Figure~\ref{fig:difficulty}(a) and~\ref{fig:difficulty}(b) clearly demonstrates a negative correlation between disease count and procedure count with patient disease popularity. This suggests that as the rarity of a patient's disease increases, their clinical condition becomes more complex. Consequently, patients with rare diseases exhibit more intricate combinations of clinical conditions.

\textbf{Observation 2}: Figure~\ref{fig:difficulty}(c) and~\ref{fig:difficulty}(d) demonstrate that patients with rarer diseases tend to have a higher number of prescribed medications and a lower popularity of these medications. This indicates that rare diseases patients require more specifically tailored medication options, resulting in more complex therapeutic regimens.

In summary, our analysis highlights the increased complexity of both clinical conditions and therapeutic regimens in patients with rarer diseases. These findings underscore the need for more advanced techniques, such as expressive encoder and pre-training techniques, to effectively address the unique challenges faced by rare disease patients.
\section{Our Method}
\begin{figure}[t]
    \centering
    \includegraphics[width=0.5\textwidth]{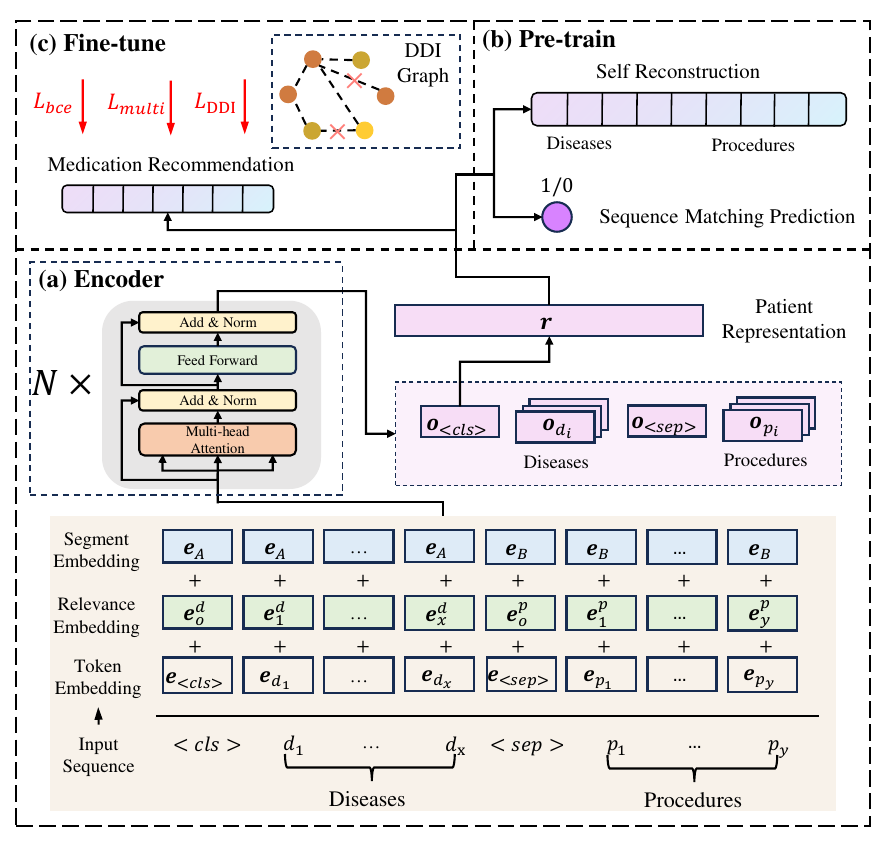}
    \vspace{-0.4cm}
    \caption{The illustration of our proposed RAREMed. We first model the patient clinical information with three embedding layers and a transformer encoder to get the patient representation. Then we pre-train our model on Sequence Matching Prediction task and Self Reconstruction task. Finally, the model is fine-tuned on medication recommendation task to get the recommended medication combination.}
    \label{fig:model}
    \vspace{-0.4cm}
\end{figure}

The model framework of our RAREMed is illustrated in Figure~\ref{fig:model}. We will commence by providing a detailed description of our encoder, which is specifically designed to effectively model patient clinical information. 
Subsequently, we will delve into the utilization the pre-training techniques to enhance the representation of input clinical codes, thereby ensuring equitable recommendation for patients with both common and rare diseases. Lastly, we will elaborate on the fine-tuning process of the pretrained model, customized for the medication recommendation task.

\subsection{Patient Representation}
In the context of medication recommendation, it is crucial to obtain a comprehensive representation of a patient's clinical condition. Traditional approaches commonly utilize disease and procedure codes to represent patient information. However, these methods have limitations. Some overlook certain codes~\cite{leap, g-bert}, while others treat them as distinct elements and simply concatenate them without considering their intricate associations~\cite{gamenet, safedrug, cognet, micron, molerec}. Consequently, these approaches result in models that are less expressive in capturing the complexity of patient data.

To address this limitation, our approach considers disease and procedure codes as a unified sequence and utilizes a transformer encoder to generate a comprehensive representation of the patient's clinical condition. Formally, for a patient $\mathcal{V} = \{\mathbf{d}, \mathbf{p}, \mathbf{m}\}$, we construct the input sequence as follows:
\begin{align}
\label{eq:1}
input = [\text{CLS}] \oplus \mathbf{d} \oplus [\text{SEP}] \oplus \mathbf{p},
\end{align}
\noindent where [CLS] represents a special token typically placed at the beginning of the sequence, and its representation can be utilized as the patient's overall representation. [SEP] indicates another special token that signifies the separation between the disease and procedure code sequences. The symbol $\oplus$ denotes the concatenation operation between sequences.

To augment the standard token embedding layer, we incorporate two additional embeddings: segment embedding and relevance embedding. The segment embedding serves to differentiate between the two categories of input codes, specifically disease and procedure, allowing the model to discern context and distinguish between various types of medical information. 
While the relevance embedding is utilized to capture the varying significance of different diseases and procedures. Rather than treating all input codes equally, we recognize that certain diseases and procedures may exert greater influence on the patient's clinical condition. To address this, we sort the input codes based on their relevance to the patient. 
Meanwhile, we introduce two learnable relevance embedding matrices to capture the priority information, in which $\mathbf{e}^d_i$ and  $\mathbf{e}^p_j$ represent the relevance embedding of the i-th disease and j-th procedure codes, respectively.

Finally, the transformer encoder processes the embedded input sequence, yielding the final patient representation $\mathbf{r}$:
\begin{align}
\mathbf{r} = \text{Encoder}(E_{\text{tok}}(input) + E_{\text{seg}}(input) + E_{\text{rel}}(input))[0],
\end{align}
\noindent where $E_{\text{tok}}$, $E_{\text{seg}}$, and $E_{\text{rel}}$ represent the token, segment, and relevance embedding layers respectively. We take the representation of [CLS] token as the patient representation.

\subsection{Pre-training}
In order to enhance the representation of clinical codes, especially for rare ones, we introduce two self-supervised pre-training tasks for the patient encoder. These tasks are specifically designed to leverage the inherent patterns and relationships within the clinical codes, enabling the model to better capture the nuances and complexities of various clinical conditions. Through pre-training on these tasks, our goal is to facilitate the acquisition of contextualized representations that can more effectively capture the intricacies of diverse clinical conditions, including rare cases.

\noindent\textbf{Task \#1: Sequence Matching Prediction (SMP):} 
The objective of this task is to capture the intricate associations between disease and procedure codes more effectively. Specifically, we aim to train the model to discern whether the disease and procedure sequences belong to the same patient or not. This task is crucial for RAREMed to better understand the contextual dependencies between different clinical codes and improve its ability to capture the underlying connections within the patient's clinical condition.

To achieve this objective, we create an unpaired example for each input sequence pair $(\mathbf{d}_i, \mathbf{p}_i)$ by randomly substituting either the disease $\mathbf{d}_i$ or procedure sequence $\mathbf{p}_i$ with a corresponding sample from a different patient. Subsequently, our model is trained using Binary Cross-Entropy (BCE) loss to distinguish between the paired and unpaired inputs:
\begin{align}
L_{SMP} = - \log(\hat{y}_i) + \log(1-\hat{y}_j),
\end{align}
\noindent where $\hat{y}_i=\sigma(W_1\mathbf{r}_i + b_1)\in\mathcal{R}$ represents the predicted probability for the paired input and $\hat{y}_j$ for the unpaired input. Here, $\sigma$ denotes the sigmoid function. $W_1\in\mathbb{R}^{dim}$ and $b_1\in\mathbb{R}$ are trainable parameters. 

\noindent\textbf{Task \#2: Self Reconstruction (SR):}
Considering the tailored medication needs in this domain, it is essential for the patient representation to retain knowledge of all the components in the input sequence, especially rare clinical codes. Specifically, in this task, we train RAREMed to reconstruct the input clinical code sequence from the modeled patient representation $\mathbf{r}$. This encourages RAREMed to capture and preserve the essential information within the clinical codes, ensuring a comprehensive representation of the patient's clinical condition.

The reconstruction loss, denoted by $L_{SR}$, is defined as:
\begin{align}
L_{SR} =& -\sum_{j=1}^{|\mathcal{D}|+|\mathcal{P}|} \left[ \mathbf{c}_j \log(\hat{\mathbf{c}}_j) + (1-\mathbf{c}_j) \log(1-\hat{\mathbf{c}}_j) \right],
\end{align}
\noindent where $\hat{\mathbf{c}} = \sigma(W_2\mathbf{r}+b_2)\in[0,1]^{|\mathcal{D}|+|\mathcal{P}|}$ represents the probabilities of all diseases and procedures reconstructed by our model,  $W_2\in\mathbb{R}^{(|\mathcal{D}|+|\mathcal{P}|)\times dim}$ and $b_2\in\mathbb{R}^{|\mathcal{D}|+|\mathcal{P}|}$ are learnable parameters. Here, $\mathbf{c}\in\{0,1\}^{|\mathcal{D}|+|\mathcal{P}|}$ represents the ground truth labels. $\mathbf{c}_j$ is set to 1 only if the corresponding label is present in the input sequence.

\subsection{Fine-tune and Inference}
After pre-training RAREMed on two tasks, we fine-tune the model to achieve accurate and fair medication recommendations. To predict the medication, we integrate a multi-label classification layer and utilize the patient representation as input:
\begin{align}
    \hat{\mathbf{o}} = \sigma(W_3 \mathbf{r} + b_3),
\end{align}
\noindent where $\hat{\mathbf{o}}\in[0,1]^{|\mathcal{M}|}$ 
is the probability of medications being recommended.
$W_3\in\mathbb{R}^{|\mathcal{M}|\times dim}$ and $b_3\in\mathbb{R}^{|\mathcal{M}|}$ are learnable parameters.

Following previous research works~\cite{safedrug, micron, drugrec, gamenet}, we fine-tune the entire model using the subsequent objective functions.
First, we treat the prediction of each medication as an independent task and use the BCE loss for optimization:
\begin{align}
L_{bce} =&-\sum_{i=1}^{|\mathcal{M}|} \left[ \mathbf{m}_i \log(\hat{\mathbf{o}}_i) + (1-\mathbf{m}_i) \log(1-\hat{\mathbf{o}}_i) \right].
\end{align}

Additionally, we employ the multi-label margin loss to ensure 
the model assigns higher scores to the correct medications compared to the incorrect ones:
\begin{align}
L_{multi} =& \sum_{i,j: \mathbf{m}_i=1, \mathbf{m}_j=0} \frac{\text{max}(0, 1-(\hat{\mathbf{o}}_i-\hat{\mathbf{o}}_j))}{|\mathcal{M}|}.
\end{align}

To ensure medication safety, we use the DDI loss to penalize drug pairs with adverse interactions as described in~\cite{gamenet, safedrug}:
\begin{align}
L_{ddi} =& \sum_{i=1}^{|\mathcal{M}|} \sum_{j=1}^{|\mathcal{M}|} \mathbf{A}_{ij} \cdot \hat{\mathbf{o}}_i \cdot \hat{\mathbf{o}}_j.
\end{align}

The balance between accuracy and safety is achieved by combining the losses using a weighted sum:
\begin{align}
L =& (1-\beta)((1-\alpha) L_{bce} + \alpha L_{multi}) + \beta L_{ddi},
\end{align}
where $\alpha$ and $\beta$ are hyperparameters balancing the contributions of the different losses.

During inference, medications with a probability greater than the threshold $\delta = 0.5$ are recommended to the patient. 
Therefore, the final set of recommended medications $\hat{\mathcal{Y}}$ can be defined as:
\begin{align}
\hat{\mathcal{Y}} = \{i | \hat{\mathbf{o}}_i > 0.5, 1\leq i \leq |\mathcal{M}|\}.
\end{align}

\section{Experiments}
In this section, we conduct comprehensive experiments to answer the following four questions:

\begin{itemize}[leftmargin=*]
    \item \textbf{RQ1}: How does the performance of the proposed RAREMed compare to that of existing medication recommendation methods?
    \item \textbf{RQ2}: Does the RAREMed effectively mitigate unfairness in medication recommendations?
    \item \textbf{RQ3}: How do the individual components of the RAREMed influence its performance with respect to accuracy and fairness?
    \item \textbf{RQ4}: What are the influential factors, e.g., the DDI Graph, that markedly impact recommendation performance of RAREMed?
\end{itemize}

\subsection{Experimental Setup}
\noindent\textbf{Datasets.}
We used EHR data from two authentic EHR datasets, namely MIMIC-III~\cite{mimic-iii}, MIMIC-IV~\cite{mimic-iv} and DDI data extracted from the TWOSIDES database~\cite{ddi}. Following previous works~\cite{gamenet, safedrug, drugrec}, we processed the datasets and randomly divided them into training, validation, and testing sets in a ratio of 4:1:1.
The statistics of processed datasets are detailed in Table~\ref{tab:statistics}.

\noindent\textbf{Evaluation Protocol.} \label{group}
We evaluate the overall performance of all methods using widely accepted metrics~\cite{gamenet, safedrug, cognet, drugrec, molerec}, including Jaccard coefficient (Jaccard), Precision-Recall Area Under Curve (PRAUC), F1-score (F1), DDI rate (DDI), and the average number of recommended medicines (\#MED), providing a comprehensive assessment of overall recommendation performance. 
It is worth noting that higher values of Jaccard, PRAUC, and F1 indicate improved accuracy, while a lower DDI value suggests enhanced medication safety. Furthermore, we believe that a successful medication recommender should resemble the behavior of doctors by recommending a similar number of medications, as reflected in the metric \#MED.

Details of metrics are listed below. 
Building upon the formulation detailed in Section~\ref{sec:formulation}, $\mathcal{M}$ represents the set of all medications in the dataset, while $\mathbf{m}, \hat{\mathbf{m}}\in \{0, 1\}^{|\mathcal{M}|}$ signify the prescribed and predicted medications to a patient, respectively.

\begin{align*}
    \text{Jaccard} = \frac{\{i:\mathbf{m}_i=1\}\cap\{j:\hat{\mathbf{m}}_j=1\}}{\{i:\mathbf{m}_i=1\}\cup\{j:\hat{\mathbf{m}}_j=1\}}.
\end{align*}

\begin{align*}
    \text{F}_1=\frac{2\text{R}\times \text{P}}{\text{R}+\text{P}},
\end{align*}
\noindent where the recall and precision are formulated as
\begin{align*}
    \text{R}=\frac{\{i:\mathbf{m}_i=1\}\cap\{j:\hat{\mathbf{m}}_j=1\}}{\{i:\mathbf{m}_i=1\}},
    \text{P}=\frac{\{i:\mathbf{m}_i=1\}\cap\{j:\hat{\mathbf{m}}_j=1\}}{\{j:\hat{\mathbf{m}}_j=1\}}.
\end{align*}

\begin{align*}
    \text{PRAUC}=\sum_{k=1}^{|\mathcal{M}|}\text{P}_k(\text{R}_k-\text{R}_{k-1}),
\end{align*}
\noindent where $\text{P}_k, \text{R}_k$ represent the precision and recall at cut-off $k$.

\begin{align*}
    \text{DDI}=\frac{\sum_{l,k\in\{i:\hat{\mathbf{m}}_i=1\}}\mathbf{A}_{lk}}{\sum_{l,k\in\{i:\hat{\mathbf{m}}_i=1\}}1},
\end{align*}
where $\mathbf{A}$ is the DDI graph defined in Section~\ref{sec:formulation}.

To examine the fairness dimension, we conduct a group-wise comparison by categorizing all patients into five groups, denoted as $\{\mathcal{G}_1, \cdots, \mathcal{G}_5\}$, according to the occurrence frequency of the least common disease codes they exhibit.
$\mathcal{G}_1$ represents patients with the most commonly observed clinical codes, while $\mathcal{G}_5$ comprises patients with the rarest codes. We calculate the average Jaccard coefficients for each group to assess the accuracy of recommendations for patients with different code occurrences. Additionally, we compute the standard deviation (denoted as $\sigma$) of these five Jaccard coefficients to evaluate whether the models perform similarly across various groups, indicating fairness in the recommendations.

\begin{table}[t]
\centering
\caption{Statistics of processed data.}
\label{tab:statistics}
\vspace{-0.2cm}
\begin{tabular}{l|rr}
\hline
\textbf{Item}  & \textbf{MIMIC-III} &  \textbf{MIMIC-IV} \\
\hline
\# of visits / \# of patients    & 14949/6344   & 16117/6352    \\
dis. / proc. / med. space size  & 2026/630/112 & 3415/1072/123 \\
avg. / max \# of visits          & 4.96/29      & 7.30/47       \\
avg. / max \# of dis. per visit & 13.76/39     & 12.66/39           \\
avg. / max \# of proc. per visit & 4.45/28      & 2.53/28       \\
avg. / max \# of med. per visit  & 19.60/52      & 12.79/53        \\
\hline
\end{tabular}
\vspace{-0.8cm}
\end{table}

\begin{table*}[t]
\centering
\caption{Overall performance comparison in the single-visit setting. RETAIN and G-Bert are excluded as they cannot generate recommendations for single-visit patients. Note that DDI rates of ground truth on MIMIC-III and MIMIC-IV are 0.082 and 0.081 respectively. The best and the runner-up results in each column are highlighted in bold and underlined, respectively.}
\vspace{-0.2cm}
\label{tab:single}
\begin{tabular}{l|rrrrr|rrrrr}
\hline
 & \multicolumn{5}{c|}{MIMIC-III} & \multicolumn{5}{c}{MIMIC-IV} \\ \hline
Method & \multicolumn{1}{c}{Jaccard} & \multicolumn{1}{c}{PRAUC} & \multicolumn{1}{c}{F1} & \multicolumn{1}{c}{DDI} & \multicolumn{1}{c|}{\#MED} & \multicolumn{1}{c}{Jaccard} & \multicolumn{1}{c}{PRAUC} & \multicolumn{1}{c}{F1} & \multicolumn{1}{c}{DDI} & \multicolumn{1}{c}{\#MED} \\ \hline
LR       & 0.4933 & 0.7639 & 0.6519 & 0.0786 & 16.85 & 0.4150 & 0.6782 & 0.5650 & 0.0736 & 9.85   \\
LEAP     & 0.4526 & 0.6583 & 0.6154 & 0.0722 & 19.04 & 0.3907 & 0.5540 & 0.5437 & 0.0548 & 12.66  \\
GAMENet  & 0.5210 & 0.7780 & 0.6762 & 0.0781 & 19.78 & 0.4401 & 0.6833 & 0.5933 & 0.0718 & 12.89  \\
SafeDrug & 0.5255 & 0.7732 & 0.6804 & 0.0688 & 20.84 & {\ul 0.4560} & 0.6858 & {\ul 0.6098} & 0.0689 & 14.07  \\
COGNet   & 0.5109 & 0.6465 & 0.6676 & 0.0737 & 25.05 & 0.4313 & 0.6112 & 0.5850 & 0.0866 & 15.46  \\
MICRON   & 0.5119 & 0.5190 & 0.6676 & {\ul 0.0610} & 20.94 & 0.4495 & 0.4353 & 0.6033 & \textbf{0.0502} & 14.50  \\
MoleRec  & {\ul 0.5303} & {\ul 0.7795} & {\ul 0.6844} & 0.0692 & 21.30 & 0.4502 & {\ul 0.6867} & 0.6040 & 0.0699 & 14.05  \\\hline
\textbf{RAREMed}  & \textbf{0.5414} & \textbf{0.7922} & \textbf{0.6942} & \textbf{0.0529} & 19.66 & \textbf{0.4625} & \textbf{0.7008} & \textbf{0.6158} & {\ul 0.0508} & 12.51  \\\hline
\end{tabular}
\end{table*}

\begin{table*}[t]
\centering
\caption{Overall performance comparison under the multi-visit setting.}
\vspace{-0.2cm}
\label{tab:multi}
\begin{tabular}{l|rrrrr|rrrrr}
\hline
 & \multicolumn{5}{c|}{MIMIC-III} & \multicolumn{5}{c}{MIMIC-IV} \\ \hline
Method & \multicolumn{1}{c}{Jaccard} & \multicolumn{1}{c}{PRAUC} & \multicolumn{1}{c}{F1} & \multicolumn{1}{c}{DDI} & \multicolumn{1}{c|}{\#MED} & \multicolumn{1}{c}{Jaccard} & \multicolumn{1}{c}{PRAUC} & \multicolumn{1}{c}{F1} & \multicolumn{1}{c}{DDI} & \multicolumn{1}{c}{\#MED} \\ \hline
LR       & 0.4933 &  0.7639 & 0.6519 & 0.0786 & 16.85 & 0.4150 & 0.6782 & 0.5650 & 0.0736 & 9.85 \\
LEAP     & 0.4526 &  0.6583 & 0.6154 & 0.0722 & 19.04 & 0.3907 & 0.5540 & 0.5437 & 0.0548 & 12.66 \\
RETAIN   & 0.4922 &  0.7560 & 0.6517 & 0.0792 & 24.76 & 0.4113 & 0.6543 & 0.5674 & 0.0839 & 16.95 \\
G-Bert   & 0.5037 &  0.7631 & 0.6617 & 0.0832 & 24.46 & 0.4292 & 0.6746 & 0.5822 & 0.0794 & 16.19 \\
GAMENet  & 0.5205 &  0.7767 & 0.6754 & 0.0774 & 19.52 & 0.4462 & 0.6923 & 0.5979 & 0.0807 & 12.60 \\
SafeDrug & 0.5232 &  0.7699 & 0.6781 & 0.0698 & 20.67 & 0.4566 & 0.6867 & 0.6080 & 0.0644 & 14.00 \\
COGNet   & 0.5177 &  0.6360 & 0.6715 & 0.0757 & 24.95 & 0.4518 & 0.6359 & 0.6026 & 0.0846 & 15.70 \\
MICRON   & 0.5188 &  0.6398 & 0.6735 & {\ul 0.0658} & 19.51 & 0.4555 & 0.5350 & 0.6081 & {\ul 0.0528} & 13.24 \\ 
MoleRec  & {\ul 0.5293} &  {\ul 0.7786} & {\ul 0.6838} & 0.0698 & 21.27 & {\ul 0.4576} & {\ul 0.6933} & {\ul 0.6097} & 0.0683 & 14.19 \\\hline
\textbf{RAREMed}  & \textbf{0.5414} & \textbf{0.7922} & \textbf{0.6942} & \textbf{0.0529} & 19.66 & \textbf{0.4625} & \textbf{0.7008} & \textbf{0.6158} & \textbf{0.0508} & 12.51  \\\hline
\end{tabular}
\end{table*}

\noindent\textbf{Compared Methods.}
We compare RAREMed against the following baseline algorithms:

\begin{itemize}[leftmargin=*]
    \item \textbf{LR} is a standard logistic regression technique, 
    where inputs are represented as a multi-hot vector of length $|\mathcal{D}|+|\mathcal{P}|$.
    
    \item \textbf{LEAP} \cite{leap} is an instance-based method, which employs the LSTM model to generate medication sequence.
    
    \item \textbf{RETAIN} \cite{retain} is a longitudinal model that utilizes two-level neural attention mechanism to predict patients' future condition.
    
    \item \textbf{G-Bert} \cite{g-bert} integrates the GNN representation into transformer-based visit encoders, which is pre-trained on single-visit data.
    
    \item \textbf{GAMENet} \cite{gamenet} uses memory neural networks and graph convolutional networks to encode historical EHR data and DDI graph.
    
    \item \textbf{SafeDrug} \cite{safedrug} leverages the drug molecular graph and DDI graph to ensure the safety of medication recommendations.
    
    \item \textbf{COGNet} \cite{cognet} uses a novel copy-or-predict mechanism which frames drug recommendation as a sequence generation problem.

    \item \textbf{MICRON} \cite{micron} is an recurrent residual learning model that focuses on the change of medications.

    \item \textbf{MoleRec}~\cite{molerec} models 
    the dependencies between patient's health condition and molecular substructures.
    
\end{itemize}
In addition to the medication recommendation methods mentioned above, we also implement a baseline approach specifically designed for fair recommendation in group-wise comparison:

\begin{itemize}[leftmargin=*]
    \item \textbf{Rebalancing}: 
    We employ a resampling technique~\cite{kamiran2009classifying, ekstrand2018all} to address the issue of data imbalance. Specifically, 
    we calculate the Inverse Propensity Score (IPS) score~\cite{ipsmethod, ipsevaluator} for each disease code. Subsequently, we assign each patient a weight score based on the IPS score associated with the rarest disease. 
    In each training epoch, we perform random sampling with replacement from the training data, the probability of selecting a particular patient is proportional to their assigned patient weight. 
    We conduct experiment for this approach on our RAREMed without pre-training.
\end{itemize}

We omit the evaluation of the 4SDrug, DrugRec, and REFINE methods from our study due to their reliance on additional symptom information~\cite{4sdrug, drugrec} or unavailability of the source codes~\cite{refine}.

\noindent\textbf{Implementation Details.}
The hyper-parameters of all models are determined based on their performance on the validation set. RAREMed is pre-trained on the training set in a sequential manner, first on the SMP task and then on the SP task, for 30 epochs respectively. The transformer encoder in our model consists of 3 layers with 4 attention heads, and the embedding dimension is set to 512. The weights of the loss function, denoted as $\alpha$ and $\beta$, are set to 0.03 and 0.7, respectively. 
The parameters are trained using the AdamW optimizer~\cite{adamw} with a learning rate of 1e-5 and weight decay of 0.1. 

\subsection{Overall Performance Comparison (RQ1)}
We conduct experiments on two settings for a thorough comparison: 
(1) \textbf{Single-visit setting}, where each visit is regarded as a separate patient without historical records as described in Section~\ref{sec:formulation}; (2) \textbf{Multi-visit setting}, where a patient may have multiple visits, and the historical records are available when recommending medications for the second and subsequent visits.
This setting is designed for a fair comparison with longitudinal methods~\cite{gamenet, safedrug, cognet, micron, molerec}.

\begin{table*}[t]
\centering
\caption{Ablation study on MIMIC-IV dataset. Here, $\sigma$ represents the standard deviation of Jaccard across five groups.}
\label{tab:abla}
\begin{tabular}{l|ccccc|ccccc|c}
\hline
\multirow{2}{*}{Model} & \multicolumn{5}{c|}{Overall Performance} & \multicolumn{6}{c}{Group-wise Performance}\\
\cline{2-12}
 & Jaccard & PRAUC & F1 & DDI & \#MED & $\mathcal{G}_1$ & $\mathcal{G}_2$ & $\mathcal{G}_3$ & $\mathcal{G}_4$ & $\mathcal{G}_5$ & $\sigma$ 
 \\ \hline
\textbf{$\mathbf{w/o}$ P}    &  0.4561 & 0.6939 & 0.6092 & 0.0509 & 12.94 & 0.4719 & 0.4666 & 0.4534 & 0.4475 & 0.4410 & 0.01157\\
\textbf{$\mathbf{w/o}$ U}    &  0.4589 & 0.6965 & 0.6122 & 0.0545 & 13.13 & \textbf{0.4773} & \textbf{0.4707} & 0.4558 & 0.4485 & 0.4423 & 0.01320\\
\textbf{$\mathbf{w/o}$ s\&r} &  0.4528 & 0.6892 & 0.6062 & 0.0554 & 13.15 & 0.4695 & 0.4646 & 0.4508 & 0.4451 & 0.4342 & 0.01286\\\hline
\textbf{RAREMed} & \textbf{0.4625} & \textbf{0.7008} & \textbf{0.6158} & \textbf{0.0508} & 12.51& \textbf{0.4773} & 0.4702 & \textbf{0.4592} & \textbf{0.4544} & \textbf{0.4521} & \textbf{0.00962}\\\hline
\end{tabular}
\end{table*}

Based on the results presented in Table~\ref{tab:single} and Table~\ref{tab:multi}, we have the following observations:
\begin{itemize}[leftmargin=*]

    \item \textbf{Baseline Comparison.} LR and RETAIN, which were not originally designed for medication recommendation tasks, exhibit poor performance. LR overlooks the interaction among clinical codes, while RETAIN fails to adequately capture the current patient condition. Similarly, LEAP and G-Bert, which do not consider the procedure sequence, also underperform in this domain. 
    Furthermore, COGNet and MICRON utilize historical information in a specific manner, but their performance significantly declines in the single-visit setting. 
    By integrating external knowledge such as EHR graph and drug molecule structures, GAMENet, SafeDrug and MoleRec achieve better performance. Interestingly, these three methods perform even worse in multi-visit setting compared to single-visit setting with MIMIC-III, highlighting the complexity of leveraging historical information. 
    
    \item \textbf{Accuracy of RAREMed.} Our RAREMed surpasses all the baseline methods in terms of accuracy, exhibiting higher Jaccard coefficient, F1 score, and PRAUC on both the MIMIC-III and MIMIC-IV dataset. This superiority is observed not only in the single-visit scenario (Table~\ref{tab:single}) but also in the multi-visit scenario (Table~\ref{tab:multi}). Remarkably, RAREMed achieves this improvement without relying on historical records. This highlights the efficacy of our medication recommendation framework, which can be attributed to the meticulous design of the expressive encoder and the effectiveness of the pre-training tasks employed.
    
    \item \textbf{Security of RAREMed.} RAREMed ensures the lowest DDI rate compared to 
    the ground-truth level and 
    baseline methods, with few exceptions. This showcases the system's capability to effectively balance between accuracy and safety, a topic that will be further explored in Section~\ref{sec: params}.
    
\end{itemize}

\subsection{Group-wise Performance Comparison (RQ2)}

\begin{figure}[t]
\centering
\includegraphics[width=0.47\textwidth]{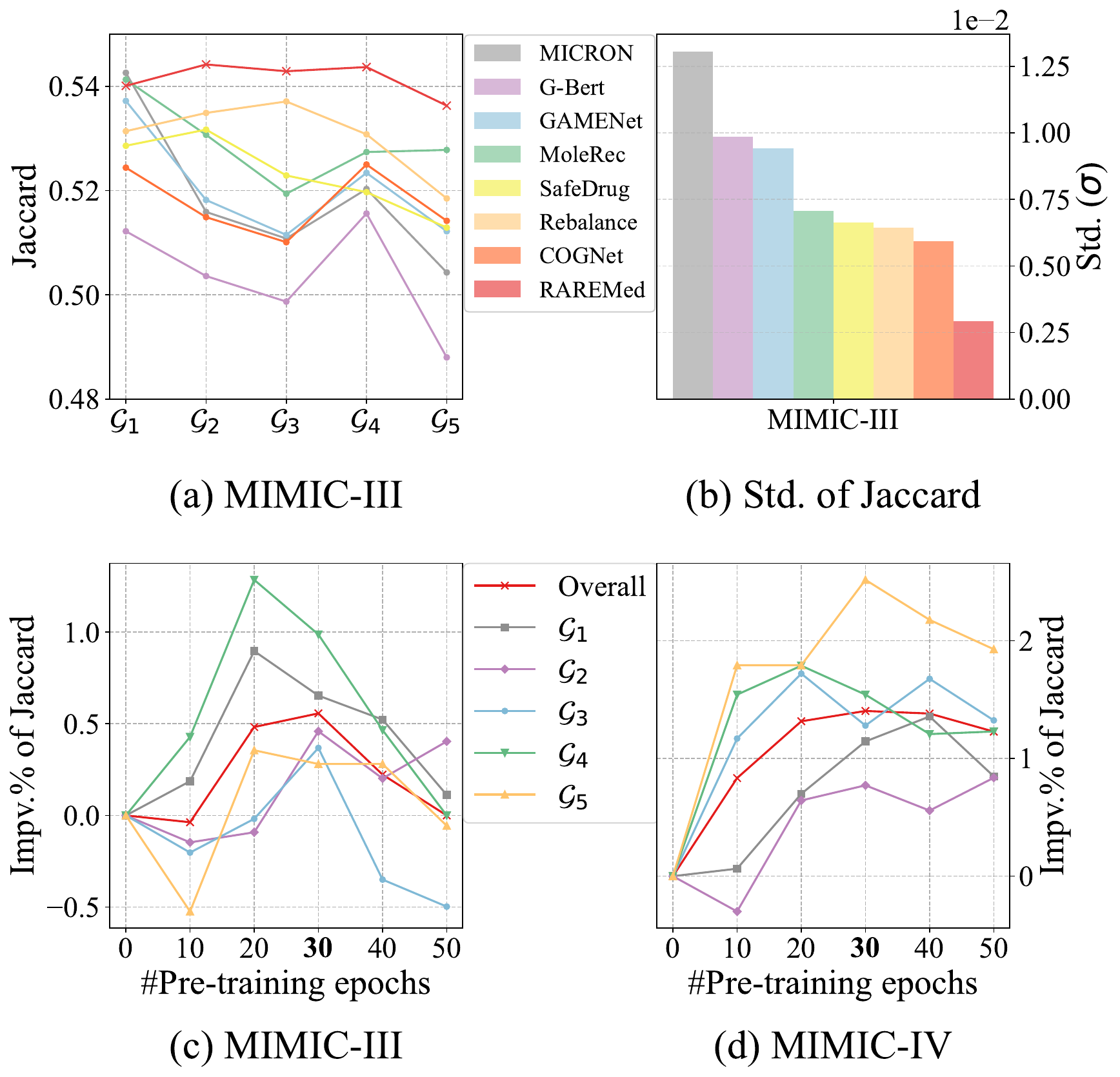}
\caption{(a) Jaccard coefficients on various patient groups. (b) Standard deviation of Jaccard coefficients across groups. (c) and (d) The improvement ratio of Jaccard across groups w.r.t. the number of pre-training epochs. The bold number indicates the selected hyper-parameter.}
\vspace{-0.4cm}
\label{fig:fairness}
\end{figure}

In this subsection, our objective is to examine model performance among patient groups with varying levels of disease prevalence.
Figure~\ref{fig:fairness} depicts the performance of RAREMed and baseline methods, from which the following observations can be made:

\begin{itemize}[leftmargin=*]
    \item \textbf{Baseline Methods Yield Unfair Recommendations.} Conventional medication recommendation methods exhibit a notable decrease in accuracy as the frequency of disease codes decreases, as illustrated in Figure~\ref{fig:fairness}(a). This leads to a higher standard deviation of Jaccard across different groups, as depicted in Figure~\ref{fig:fairness}(b). Notably, G-Bert~\cite{g-bert}, which utilizes pre-training techniques, struggles to achieve satisfactory accuracy for patients with rare diseases, potentially due to its disregard for procedure codes and subsequent failure to capture a comprehensive representation of patients' clinical information during pre-training. This underscores the superiority of our model structure and pre-training task design. Additionally, Rebalancing, tailored to address fairness concerns, also underperforms compared to RAREMed in terms of both fairness and overall performance.
    
    \item \textbf{RAREMed Effectively Addresses Fairness Concerns.} As depicted in Figure~\ref{fig:fairness}(a), RAREMed demonstrates superior performance for patients with rare diseases, while maintaining efficacy for the common group. Additionally, RAREMed showcases consistent performance across patient groups, as evidenced by minimal variance in Jaccard compared to the baseline methods. This trend highlights the strength of RAREMed in acquiring robust representations, particularly for rare clinical codes, and its capability to deliver fair medication recommendations for patients with rare diseases.
    
    \item \textbf{Pre-training Demonstrates Effectiveness on Biased Datasets.} The results presented in Figure~\ref{fig:fairness}(c) and (d) highlight the overall positive impact of our pre-training strategy on both 
    common group and rare group, with few exceptions. Specifically, on MIMIC-III, where the disease distribution is less skewed, our pre-training strategy consistently enhances performance across various groups, as evidenced by the uniform improvement ratio observed after 30 pre-training epochs. In contrast, on MIMIC-IV, characterized by a more biased disease distribution and presenting greater challenges\footnote[6]{
    This is supported by the extensive distribution of long-tail disease codes in MIMIC-IV, as illustrated in Figure~\ref{fig:intro}, and the greater variance of Jaccard scores across groups, as presented in Table~\ref{tab:abla}, in comparison to the findings in Figure~\ref{fig:fairness}(b).
    }, RAREMed notably improves the performance of the rarest group, effectively mitigating unfairness. 
    The substantial improvement in overall accuracy on such a more biased dataset is also noteworthy.
\end{itemize}

\subsection{Ablation Study (RQ3)}
To evaluate the effectiveness of each component in RAREMed, we conduct ablation study with the following ablation models:
\begin{itemize}[leftmargin=*]
    \item \textbf{``$\mathbf{w/o}$ P''}, which removes two pre-training tasks, resulting in initializing parameters of encoder and embedding layers randomly.
    \item \textbf{``$\mathbf{w/o}$ U''}, which disables the unified encoder and learns two encoders for diseases and procedures seperately and then concatenates the output. This setup retains two pre-training tasks.
    \item \textbf{``$\mathbf{w/o}$ s\&r''}, it disables \textit{segment} and \textit{relevance} embedding layers,
    and patient representation relies solely on the \textit{token} embedding.
\end{itemize}

We omit results on MIMIC-III, as they align with the same conclusion. We have the following observations from results in Table~\ref{tab:abla}:

\begin{itemize}[leftmargin=*]
    \item \textbf{``$\mathbf{w/o}$ P''.} 
    As anticipated, the diminished performance observed in the "w/o P" variant underscores the importance of our pre-training tasks. Notably, RAREMed outperforms "$w/o$ P" across all five groups, affirming that our pre-training strategy not only enhances performance on rare diseases but also preserves the efficacy of the original model on other diseases. The enhancement seen in RAREMed can be attributed to the augmented encoder obtained through the pre-training phase, enabling RAREMed to grasp intricate disease/procedure associations and produce more comprehensive patient representations.
    \item \textbf{``$\mathbf{w/o}$ U''.} RAREMed surpasses the ``$w/o$ U'' variant, particularly in terms of fairness, further emphasizing the importance of capturing and comprehending the associations between diseases and procedures for fair medication recommendations.
    \item \textbf{``$\mathbf{w/o}$ s\&r''.} RAREMed also outperforms the ``$w/o$ s\&r'' variant. Notably, the ``$w/o$ s\&r'' variant also exhibits significantly fluctuating performance across different patient groups. These results highlight the crucial role played by the segment and relevance embedding layers in capturing the complex relationships among diseases and procedures, particularly for rare ones.
\end{itemize}

\subsection{Hyperparameter Studies (RQ4)}
\label{sec: params}
We conducted an investigation to understand the influence of hyperparameters on the efficacy of RAREMed on the MIMIC-IV. Specifically, we considered four hyperparameters: the multi-label margin loss weight ($\alpha$), the DDI loss weight ($\beta$), the number of pre-training epochs (\#Epochs), and the embedding space dimension ($dim$).

\begin{figure}[t]
\centering
\includegraphics[width=0.47\textwidth]{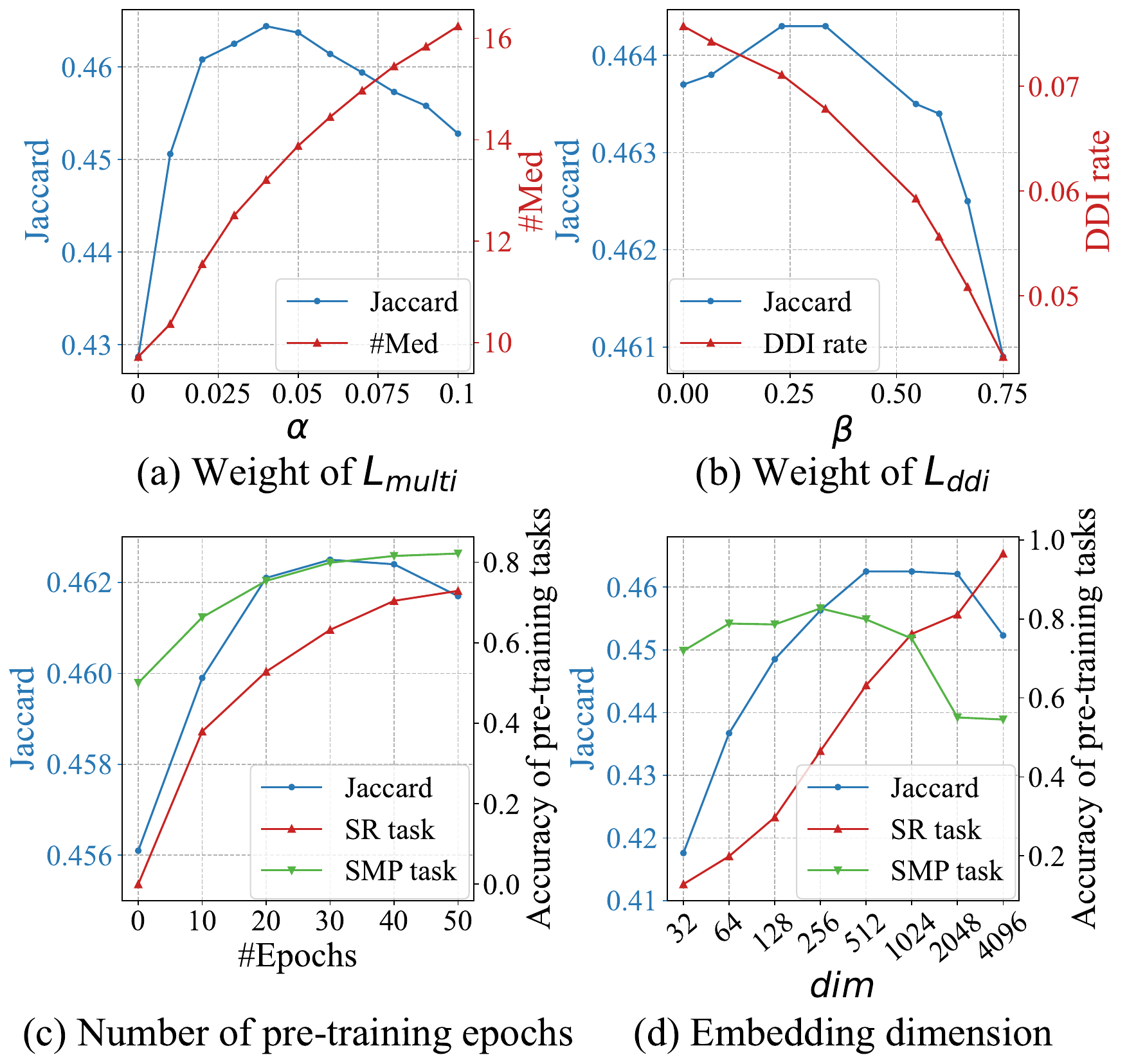}
\caption{Hyperparameter effects on model performance.}
\label{fig:params}
\end{figure}

\noindent\textbf{Loss Weight.} Figure~\ref{fig:params}(a) demonstrates that the value of $\alpha$ significantly affects the size of the recommended medicine set. Additionally, we observed that the recommendation accuracy initially increases and then declines as the number of recommended medications increases.
This suggests that in addition to the ranking task commonly considered in classic recommendation systems, the size of recommended medication set is also an important factor in the field of medication recommendation. 

On the other hand, Figure~\ref{fig:params}(b) illustrates the impact of $\beta$ on the DDI rate and Jaccard coefficient. As $\beta$ increases, the DDI rate consistently decreases, while the Jaccard increases briefly and then drops. This indicates that, overall, there is a trade-off relationship between accuracy and security, while uncontrollable DDI (when $\beta=0$) also harms accuracy. Notably, even when the DDI rate is restricted to less than 5\% (when $\beta=0.75$), the accuracy of RAREMed remains relatively high, demonstrating its robustness.

\noindent\textbf{Number of Pre-training Epochs.} Figure~\ref{fig:params}(c) illustrates the impact of the number of pre-training epochs on the pre-training tasks and the medication recommendation task. It is evident that as the number of pre-training epochs increases, the model's  performance in pre-training tasks (SR task and SMP task) improves. This observation indicates that the model acquires knowledge about clinical code associations and gains comprehensive patient representations during pre-training, which are crucial for medication recommendation, particularly for patients with rare diseases. 
However, the improvement in the downstream task (Jaccard) starts to diminish after approximately 30 epochs. This phenomenon may be attributed to overfitting in the pre-training task and the inherent differences between the pre-training and downstream tasks. These findings highlight the significance of striking an optimal balance between the performance on pre-training tasks and the downstream medication recommendation task.

\noindent\textbf{Embedding Dimension.}  As depicted in Figure~\ref{fig:params}(d), the increase in embedding dimension initially enhances RAREMed's performance on both pre-training and downstream tasks, indicating an augmented expressive capacity. However, beyond a certain threshold, the model's performance on SMP and downstream tasks begins to decline, note that RAREMed is pre-trained first on the SMP task and then on the SR task, excluding the influence of the SR task on the SMP task. This decline in performance can be attributed to training challenges such as gradient vanishing or explosion, which disrupt convergence and hinder the model's stability.

\section{Conclusion}

In this paper, we propose RAREMed, a novel medication recommendation model designed to addresses the fairness issue in medication recommendation systems. 
By focusing on enhancing accuracy for patients with rare diseases, RAREMed leverages two self-supervised pre-training tasks to learn specialized medication needs and interrelations.
The model also utilizes a unified input sequence approach to capture complex relationships among disease and procedure codes. 
Experimental results on real-world datasets demonstrate the effectiveness of RAREMed in providing accurate drug recommendations for both rare and common disease patients, thereby mitigating unfairness.

In future research, we intend to further explore the utilization of patients' historical records and additional clinical information, such as demographics, lab data, notes, and imaging, to improve the accuracy of medication recommendations. 
Additionally, incorporating more external knowledge during the pre-training phase and exploring transfer learning across different electronic health record (EHR) datasets are also promising avenues for further development. 

\begin{acks}
This work is supported by the National Natural Science Foundation of China (62272437, 62302321) and the University Synergy Innovation Program of Anhui Province (GXXT-2023-071).
\end{acks}

\clearpage
\newpage

\bibliographystyle{ACM-Reference-Format}
\balance
\bibliography{ref}

\clearpage

\end{document}